ARTICLE

# The Benchmarking Epistemology: Validity Theory for Evaluating Machine Learning Models

Timo Freiesleben[1] and Sebastian Zezulka[2]

[1]Munich Center for Mathematical Philosophy, LMU Munich, Germany and [2]Cluster of Excellence: "Machine Learning: New Perspectives for Science", University of Tübingen, Tübingen, Germany
**Corresponding author:** Sebastian Zezulka; Email: sebastian.zezulka@uni-tuebingen.de

(Received 28 October 2025 UTC; revised 07 August 2026 UTC; accepted 10 August 2026 UTC)

## Abstract

Predictive benchmarking, evaluating machine learning models based on predictive performance and competitive ranking, is central to machine learning research and scientific inquiry. However, benchmark scores at best measure performance relative to a specific dataset and learning problem. Drawing substantial scientific inferences requires additional assumptions. Adapting ideas from psychological validity theory, we propose validity conditions that make these assumptions explicit. In two case studies—ImageNet and the Fragile Families Challenge—we show how benchmark results can support inferences about research progress and limits of predictability, situating predictive benchmarking as a distinct epistemic practice in machine learning.

## 1. Introduction

Predictive benchmarking is the central evaluation strategy in machine learning. By now, scientists use benchmarks to address diverse scientific questions: Do we make progress on image classification? Are life outcomes unpredictable? Do large language models (LLMs) exhibit general capabilities? This raises epistemological questions about what predictive benchmarks measure and how we can appraise the validity of inferences drawn from such measurement. To answer these questions, we begin by tracing the historical development of predictive benchmarking, from a methodology for evaluating machine learning systems to an epistemic tool used across the sciences to justify claims that benchmarks were not originally designed to support.

In the early days of artificial intelligence research, the Defense Advanced Research Projects Agency (DARPA) was one of its main funders. Much like today, researchers promised to solve complex and abstract engineering tasks like language translation or object recognition within just a few months or years, ultimately never coming close to delivering on their promises. A revealing episode played out in 1966, when Minsky and Papert, whose research was strongly supported by DARPA at the time, assigned a group of students a project aimed at artificially constructing "a significant portion of a visual system" over the course of just one summer (Papert 1966). Following substantial cuts in government funding and disappointed expectations, DARPA employees were faced with the difficult question of assessing whether progress had been made on any of the problems.

DARPA answered this question in the late 1980s by adopting the *Common Task Framework* (Liberman 2010; Donoho 2017). Instead of the grand claims about solving artificial general intelligence that dominated the early days of artificial intelligence (AI), the Common Task Framework shifted the focus to concrete (usually supervised) learning problems. Data for building models to solve these problems was made publicly available, and crucially, performance was evaluated on unseen holdout data. DARPA had both the authority and the resources to establish the Common Task Framework as the new norm in AI research, thereby laying the groundwork for modern predictive benchmarking and fostering the shift from symbolic AI toward machine learning.[1]

A *predictive benchmark* has four components: (1) a *learning problem* specifying the prediction target; (2) a standardized *dataset* split into a training set and an evaluation set; (3) an *evaluation metric*; and (4) a public *leaderboard*, ranking the models by the scores they obtained.[2] Take the famous Modified National Institute of Standards and Technology (MNIST) benchmark as an example (Lecun et al. 1998). It consists of a seemingly simple learning problem: Given a small grayscale image of a handwritten digit, identify which digit it represents, from 0 to 9 (Yadav and Bottou 2019). Performance of models is evaluated by their accuracy on the evaluation set. For some years, MNIST was a proving ground for new machine learning techniques, and a public leaderboard still tracks model results. But times have changed, and with current methods reaching about 99.87% accuracy, the benchmark is considered too easy and essentially "solved."

Predictive benchmarks have since evolved from a tool for making funding decisions into the dominant methodology for evaluating machine learning models and coordinating research programs around shared problems.[3] They have become the "iron rule" of machine learning (Hardt 2026): Justification for algorithmic innovation largely depends on outperforming the "state of the art," as measured by a relevant benchmark. With the growing influence of machine learning over the last decade, predictive benchmarks have transcended their engineering roots and are emerging as a relevant methodology in the empirical sciences. This became very clear in 2024, when benchmark-driven research earned its first Nobel Prize: John Jumper and Demis Hassabis were honored for their protein structure prediction models, which substantially outperformed all previous methods on the Critical Assessment of

[1] The Common Task Framework echoed ideas introduced as early as 1959, when Bill Highleyman outlined many of the principles in his PhD thesis on handwritten character recognition (Orr and Crawford 2024). Highleyman also shared his dataset with other researchers—by mail.

[2] Recent benchmarks like Super General Language Understanding Evaluation (SuperGLUE) (Wang et al. 2019) or Holistic Evaluation of Language Models (HELM) (Bommasani et al. 2023) often encompass several learning problems, datasets, and metrics. This raises additional questions about the aggregation of scores (Zhang and Hardt 2024).

[3] Hardt and Recht (2022), chap. 8, Paullada et al. (2021), and Koch and Peterson (2024) discuss the historical development of predictive benchmarks.

Protein Structure Prediction (CASP)[4] benchmark challenge (Jumper et al. 2021). And it is not only structural biology—prominent predictive benchmarks now exist across many scientific fields, including meteorology (Rasp et al. 2020), medicine (Irvin et al. 2019), neuroscience (Schrimpf et al. 2018), materials science (Dunn et al. 2020), legal studies (Guha et al. 2023), psychology (Fried et al. 2023), and sociology (Sivak et al. 2024).

Despite the central role that benchmarks play in machine learning and in the sciences, they have received little attention from philosophers. Philosophers of machine learning have primarily concentrated on issues of epistemic opacity (Creel 2020; Boge 2021; Sullivan 2022; Duede 2023) and explainable AI (Zednik and Boelsen 2022; Watson 2022; Buchholz and Grote 2023; Freiesleben et al. 2024). A rich philosophical literature examines (big) data along its representational and relational dimensions (Bogen and Woodward 1988; Leonelli 2015; Leonelli 2020), but not in the context of benchmarking. Formal epistemologists, providing a priori guarantees for learning algorithms, have typically adopted a theoretical perspective (Herrmann 2020; Sterkenburg and Grünwald 2021; Lin 2022; Grote et al. 2024b; Sterkenburg 2024), largely overlooking the empirically driven methodology of predictive benchmarking that underpins much of contemporary machine learning research. Yet it is precisely because benchmarking structures entire research programs and informs scientific inquiry and policy decisions that it demands philosophical scrutiny. Without an epistemic basis, researchers risk drawing misleading or irreproducible inferences from benchmark scores. This article contributes such a basis, showing how predictive benchmarks can serve as epistemic tools capable of supporting a wide range of scientific inferences, if researchers explicitly consider the relevant *validity conditions*. Specifically, we address the following question: How can benchmark scores support substantive scientific inferences within machine learning and across the sciences?

We argue that, analogous to psychometric and educational tests, predictive benchmarks are best understood as measurement tools (sec. 2). The resulting measurements $M(F)$ warrant a *generalization inference* about the expected predictive performance of the set of models $F$ on a given learning problem, given appropriate sampling conditions. Drawing more substantial inferences $I$ from the measurements requires additional assumptions $A$ to render the inferences valid, that is,

$$M(F) \wedge A \rightarrow I.$$

The study of these assumptions is often referred to as *validity theory* (Kane 2013; American Educational Research Association et al. 2014). We adapt the *argument-based* validity framework to the context of predictive benchmarks, comprising four steps: (1) defining the intended inference, (2) specifying necessary validity conditions for this inference, (3) providing evidence for and against the conditions, and (4) constraining the inference accordingly. We specify validity conditions under which substantial scientific inferences can be drawn from benchmark scores, establishing predictive benchmarks as key sites of conceptual and theoretical reasoning in machine learning and the sciences more broadly (sec. 3).

---

[4] The CASP challenge, dating back to 1994, is among the oldest scientific benchmark challenges (Donoho 2024).

We illustrate the framework with two case studies that reflect different stakeholders, the distinct inferences they seek to draw from benchmark scores, and the corresponding validity concerns that arise. The first examines ImageNet for computer vision, where the learning problem is to map images to predefined classes (sec. 4). Within the field, ImageNet serves as a proxy for measuring methodological progress on the broader task of image classification. For this role to be justified, in addition to a generalization inference, one must also draw an *extrapolation inference* from performance on ImageNet to performance on other such image-classification problems, and a *conceptualization inference* from performance on ImageNet to image classification more generally. We analyze the conditions under which these inferences are valid and the types of evidence required to support them.

The second case study considers the Fragile Families Challenge (FFC) from sociology, which focuses on predicting life outcomes of young people (sec. 5). The uniformly low predictive performance of models in this setting has led researchers to question whether such outcomes are fundamentally unpredictable. In addition to the preceding sub-inference, justifying this inference requires establishing additional *predictability conditions* that rule out alternative explanations, including limited sample size or unobserved features. In section 6, we conclude by outlining the broader relevance of validity theory for evaluating machine learning systems, in particular as a tool for assessing the capabilities of LLMs and the limits of benchmarking.

## 2. Predictive benchmarks are measurement tools

Predictive benchmarks are best understood as measurement instruments for the predictive performance of prediction models on specific learning problems. As such, they are analogous to psychological and educational tests that measure latent psychological traits of test takers (Blili-Hamelin and Hancox-Li 2023; Cardoso et al. 2020).[5] Consider the example of intelligence tests: Human *test takers* solve a battery of *test items*, such as sequence completion problems, to obtain a *test score*. This score is then interpreted as indicative of a trait of the test takers, conceptualized as *constructs* like their general cognitive ability. Subsequently, it may be used to draw further inferences or make decisions, such as selecting job candidates (Glymour 1998; Cave 2020). In the case of predictive benchmarking, prediction models (test takers) are evaluated on data instances in the benchmark (test items), yielding a performance score (test score). From this score, researchers draw various inferences, for example, about the predictive performance of models (constructs) to select the most suitable prediction models.

In psychological and educational measurement, well-developed methodological frameworks are used to justify inferences from test scores, commonly subsumed under the term *validity theory* (American Educational Research Association et al. 2014). They require the specification of validity conditions that warrant the inferences (Messick 1989; Kane 2013). Depending on the construct of interest, they are often complemented by a theoretical characterization of that construct (Cronbach and Meehl 1955; Borsboom et al. 2003; Millsap 2012). One expects similar efforts when it comes to validating the inferences drawn from benchmark scores.

[5] Mussgnug (2022) analyzes the output of machine learning models as measurements, rather than predictive benchmarks.

The validity conditions for a *generalization inference* from benchmark scores are well established; this is arguably the standard form of inference in machine learning. Benchmark scores are used to infer a model's performance on randomly sampled and previously unseen data. In this setting, little explicit measurement theory is required, because the problem can be formulated as a statistical estimation task: The empirical error is an estimator for the expected error. This procedure is commonly referred to as the *holdout method* (Grote et al. 2024b; Hardt 2026).

Let's unpack this inference. Evaluating a set of models on a benchmark yields numerical scores, the *empirical error* as defined by the evaluation metric.[6] For a prediction model $f$, an evaluation dataset $D_e$ with features $x_i$ and labels $y_i$, and an evaluation metric $\ell$, the empirical error is defined as

$$\hat{R}(f) := \frac{1}{\mid D_e \mid} \sum_{(x_i, y_i) \in D_e} \ell\big(f(x_i), y_i\big).$$

That is, the evaluation metric $\ell$ is used to compute an error for each instance $(x_i, y_i)$ by measuring the discrepancy between the model prediction $f(x_i)$ and the corresponding label $y_i$. These instance-level errors are then averaged over all examples in the evaluation dataset to obtain the empirical error.

To *interpret* the scores, we must connect them to the theoretical quantity or construct that the scores are intended to measure. In the case of a generalization inference, this is the *expected error*. For a model $f$ and an evaluation metric $\ell$, the expected error is defined as

$$R(f) := \mathbb{E}_{X,Y}[\ell(f(X), Y)],$$

where $(X,Y)$ are random variables drawn from a fixed but unknown data distribution, from which the evaluation data $D_e$ are assumed to be sampled.[7] The expected error captures how the models would, on average, perform on new instances from this population, effectively representing the model's performance were we able to observe an infinite number of data points.

Inferring the expected errors from benchmark scores requires supporting assumptions—we will call them *generalization conditions*—that connect them. The validity conditions must provide a sound inferential bridge between the benchmark scores and the intended interpretation. In this case, sufficient conditions are as follows:

(i) The evaluation data are independent of the models, implying, for example, that one cannot use the evaluation data for model training.

(ii) The evaluation data are statistically independent samples $(x_i, y_i)$ drawn from one identical distribution over $(X, Y)$, the independent and identically distributed (i.i.d.) assumption.

(iii) The evaluation data are sufficiently large so that the empirical error provides a reasonable approximation of the expected error.

[6] The evaluation metric does not need to coincide with the loss function used to train the model, nor must the model be trained for the learning problem on which it is evaluated.

[7] For a discussion on the interpretation of statistical assumptions, see Zhao (2025).

Together, these conditions justify the generalization inference via the law of large numbers and provide theoretical error bounds for finite samples (Hardt 2026). However, the three conditions are not strictly necessary: Weaker variants may suffice, depending on the specific requirements of the inference. Condition (i) can be weakened, because some dependencies between test data and models, such as adaptivity, do not always invalidate the inference from test error to expected error (Hardt 2026, chap. 5). However, the corresponding error bounds for finite samples, which capture the expected discrepancy between test and true error, may become substantially wider, increasing the required sample size for reliable inference (Dwork et al. 2015). Likewise, condition (ii) can be relaxed so that the test data are a random sample from an underlying population (Höltgen and Williamson 2026) or, particularly in the case of time series, to assumptions like stationarity or ergodicity (Andrews 1988; Sandrić 2017). Similarly, the requirements of condition (iii) depend on the specific inference considered: Whereas some require tight error bounds and thus large sample sizes, others can be adequately supported by smaller samples.

In any case, when the generalization conditions are warranted, they allow us to interpret benchmark scores as an implementation of the holdout method and the resulting empirical errors as estimates of the expected errors. This is a strong result, and it already enables us to draw a range of important inferences from benchmark scores: about a model's expected error on the learning problem; about how different models compare in their expected error on this problem; and even about whether particular learning algorithms, architectural choices, or hyperparameter settings yield models with lower expected errors. Generalization inferences are the basis and constitute an important step for many scientific inferences. In the words of Kane (2013), "generalizability is a necessary condition for the validity of almost all test score interpretations because almost all of these interpretations involve generalization over some universe of generalization that goes beyond the observations actually made" (29).

Although generalization inferences are relevant, in the empirical sciences and in machine learning, we often seek to draw more involved inferences from benchmark scores, which the three conditions noted previously, by themselves, cannot support. Recall our motivating examples: Researchers may wish not merely to infer that models perform well on a benchmark such as ImageNet, but that they are making progress on the broader task of image classification, or that consistently poor benchmark performance across many prediction models indicates principled limitations on the predictability of life outcomes. These inferences extend beyond statistical standards of estimation and call for a richer structure of validity considerations than is currently in place for benchmarking, potentially changing the target of estimation itself. In psychological testing, such conceptual, theoretical, and normative considerations are systematically organized under the umbrella of *validity theory* (American Educational Research Association et al. 2014). Analyzing the inferences drawn from predictive benchmarks as measurements brings into focus the central role that supporting conditions of validity play. Validity theory provides a powerful framework for conceptualizing the diverse and substantive inferences drawn from predictive benchmarks.

## 3. Validity theory for score interpretation

Two major philosophical positions can be distinguished on what validity is all about (Alexandrova and Haybron 2016; Zhao 2023). The *correspondence account* treats validity as a feature of the measurement instrument or its outcomes, for example, by establishing a causal relationship between the measurement and the construct (Borsboom et al. 2004). In contrast, the *argument-based account* understands validity as a feature of the score interpretation or its use, rather than of the measurement instrument, focusing on the logical relationship between the measurement and the intended inference (Kane 2013). Like Hood (2009), we take these accounts as complementary rather than mutually exclusive: The correspondence account provides an ontology, and the argument-based account provides an epistemology. Because we adopt an epistemological rather than an ontological perspective on benchmarks and examine them as epistemic tools for drawing valid inferences, we follow the argument-based account of validity in this article.

The argument-based account of validity has two key components: (1) the *interpretation or use argument* and (2) the *validity argument* (Kane 2013, 9). The first specifies the intended inference from the benchmark scores and the conditions under which the inference is valid; together with the measurement, these conditions should imply the intended inference. The second examines the conditions and provides empirical and conceptual warrants for them, potentially requiring us to adapt the intended inference. The uncertainties of these warrants carry over to the validity of the inference. In practice, the validity of an intended inference is established in a four-step procedure that is widely recognized as a standard in psychological and educational measurement (American Educational Research Association et al. 2014). It is defined by the following iterative sequence of instructions, where steps 1 and 2 correspond to specifying the interpretation or use argument, and steps 3 and 4 correspond to establishing the validity argument:

1. **Define the intended inference.** The inference intended to be drawn from the benchmark scores must be made specific. This includes a specification of the relevant constructs, the scope of the inference, and the substeps of the inference.
2. **Specify validity conditions.** Validity conditions must be specified that make all substeps of the intended inference and, consequently, the intended inference valid. These secure, among others, the relevant statistical assumptions and the relevance of the operationalization of the construct.
3. **Provide evidence for and against the conditions.** Researchers must provide evidence of whether and to what degree the validity conditions hold. Here, evidence is understood broadly and may include benchmark scores for relevant subgroups, results from other studies, and conceptual or domain-specific theoretical insights.
4. **Constrain the inference.** In light of the evidence, the validity conditions and, consequently, the intended inference will only be supported to some degree. There will be evidence strengthening some and evidence undermining other assumptions. Scientists must constrain the scope and content of the intended inference accordingly and then validate the inference again using the steps outlined previously.

**Table 1.** The table summarizes the five sub-inferences discussed in this article, specifying their associated validity conditions. Here, $f$ denotes the model, $D_e$ the evaluation data, $\ell$ the evaluation metric, and $d(s)$ a decision rule based on scores $s$. $\mathcal{B}$ is a set of benchmarks, and $T$ represents a theoretical task. The arrow on the right indicates that as more theoretical inferences are drawn, an increasing number of validity conditions must be satisfied

| Inference | | Conditions |
|---|---|---|
| Generalization (sec. 2) | If $f$ has empirical error/rank $s$ on benchmark $B$, then its expected error/rank on $B$ is also $s$. | (i) $D_e$ is independent of $f$; (ii) is i.i.d. sample; (iii) is sufficiently large. |
| Extrapolation (sec. 4) | If $f$ has expected error/rank $s$ on $B$, then it also has $s$ on other relevant benchmarks $B' \in \mathcal{B}$. | (i) Robust performance across related learning problems and distributions; (ii) scores informative across metrics. |
| Conceptualization (sec. 4) | If $f$ has expected error/rank $s$ across benchmarks $\mathcal{B}$, then it has $s$ on the theoretical task $T$. | (i) Learning problem describes task; (ii) data represent task classes; (iii) $\ell$ reflects task performance. |
| Predictability (sec. 5) | If a set of models $F$ has minimal expected error $s$ on task $T$, then $s$ describes optimal performance on the task $T$. | (i) models sufficiently diverse and dataset sufficiently large; (ii) relevant features measured. |
| Decision (sec. 5) | If $f$ has performance/rank $s$ on task $T$, then decision rule $d(s)$ should be used for purpose $A$. | (i) $d(s)$ reflects utility; (ii) is fair across subgroups; (iii) has no negative systemic effects. |

more theoretical inferences ↓

Applying this framework in practice imposes substantial demands on researchers and, in most cases, is not straightforward. Even specifying the intended inference can be challenging. Although researchers often start with broad, high-level inferences, there is a trade-off: The more general the intended inference is, the stronger become the required validity conditions and the more extensive must the supporting evidence be (Kane 2013). In practice, researchers often arrive at a well-specified intended inference only by an iterative process, starting from a broad formulation and refining it over multiple passes through the framework.

A key prerequisite for applying the framework is to decompose the intended inference into a set of well-motivated substeps, which we refer to as *sub-inferences*. This decomposition is necessary because intended inferences are often complex and highly context specific. Although it may be possible to formulate validity conditions directly for a given intended inference, such conditions would typically reflect this context and would be difficult to reuse, limiting the usefulness and standardization of the framework. At the same time, many inferences share common inferential steps. For example, consider the generalization inference that underlies most scientific inferences from benchmark scores. It is possible to formulate general validity conditions that can be applied across contexts for sub-inferences of this kind. A key contribution we make is to identify a set of relevant sub-inferences and to specify corresponding validity conditions. We provide an overview in table 1. Some of the inferences and conditions, especially those about generalization, extrapolation, and decision, are inspired by Kane (2013).

Specifying the appropriate validity conditions that render any specific inference valid is an important task. In the following, we identify sufficient conditions, but as discussed in the previous section for the generalization inference, these need not be necessary. Weaker validity conditions impose less demanding requirements on the supporting evidence and may thus be preferable.

Any inference is only as strong as the evidential support for its validity conditions and can be undermined by evidence that challenges the conditions. The framework thus also provides a structured method for invalidation: Inferences can be weakened or even rejected by undermining the relevant validity conditions with new evidence. As suggested by Kane (2013), the inferential steps can be seen as spans of a bridge connecting benchmark scores, validity conditions, and conclusions. A single failing span may invalidate the inference, even if others remain intact. Accordingly, researchers should focus their efforts on identifying the weakest points in the validity argument.

In general, the validity of an inference is gradual, and its assessment evolves over time (Kane 2013). More substantial or wide-ranging inferences from benchmark scores rest on more sub-inferences and, accordingly, more validity conditions. As the theoretical implications of an inference increase, we move from generalization toward extrapolation, conceptualization, decision conditions, and beyond.

The argument-based account of validity may at first appear abstract, offering a broad perspective but limited concrete guidance for specific applications. The validity conditions we propose are general in that they apply across a wide range of settings while still allowing for the flexibility needed to accommodate contextual variation. Ultimately, the conceptual subtleties of the account in benchmarking become fully apparent only in application. In the next two sections, we therefore examine two case studies. First, we analyze an inference about research progress in the context of ImageNet, a widely used benchmark in computer vision. Second, we turn to an inference about the predictability of life outcomes based on the FFC.

## 4. ImageNet: Do we make progress on image classification?

The ImageNet Challenge[8] is one of the most influential predictive benchmarks in the history of computer vision and machine learning in general (Deng et al. 2009). It was at the center of the deep learning revolution, when the convolutional neural network AlexNet (Krizhevsky et al. 2012) outperformed the competing models by about 11 percentage points in 2012. Moreover, key algorithmic innovations in deep learning, like dropout (Srivastava et al. 2014) and batch normalization (Ioffe and Szegedy 2015), were adopted because of their effectiveness in improving performance on ImageNet.

The ImageNet Challenge, particularly its classification component, became so paradigmatic in computer vision that it effectively served as a *measure of progress*. Within 5 years, researchers were able to reduce the reported error rates from 28.2% in 2010 to 3.57% in 2015 (Nguyen et al. 2018). Why can performance gains on

[8] Known as *ILSVRC*, the ImageNet Large Scale Visual Recognition Challenge.

ImageNet be taken as evidence that models have improved on image classification generally? The four-step approach introduced earlier allows us to evaluate when this inference is valid by analyzing the underlying validity conditions.

### 4.1. The intended inference

Computer vision researchers may aim to infer that *the error rates of models on ImageNet reflect their error rates on image-classification tasks generally.* If this inference holds, improvements on ImageNet can legitimately be interpreted as advances in image classification and, by extension, in computer vision. This would also justify the community's intense concentration on ImageNet.

The scope of this inference is fairly wide and ultimately concerns the theoretical notion of an image-classification task. The construct of interest is the error rate on such image-classification tasks. Connecting the scores on ImageNet to this construct requires three sub-inferences: First, the error rates on the ImageNet evaluation dataset must reflect the error rates on ImageNet-like data (generalization inference); second, they must reflect the error rates on alternative image-classification benchmarks (extrapolation inference); and finally, they must reflect error rates in image-classification tasks generally (conceptualization inference).

### 4.2. Specifying validity conditions and providing relevant evidence

To avoid jumps between the conditions supporting the inferences and their respective justifications, we present each validity condition together with a critical discussion of the supportive and contradictory evidence.

*Generalization inference*. We discussed the three standard generalization conditions in section 2, requiring that (i) the model is independent of the evaluation data, (ii) the samples are statistically independent and drawn from an identical distribution (i.i.d.), and (iii) the evaluation dataset is sufficiently large. We now consider the relevant evidence for each in turn.

The ImageNet data were used over many years, during which researchers could adapt to prior results and successful model architectures, violating the independence of the models and the evaluation data (Blum and Hardt 2015). Recht et al. (2019) recreated ImageNet, mirroring the original data-collection strategy, to empirically study this violation by reevaluating successful models like AlexNet and ResNet on the new data. By construction, these new data are independent of the models.[9] They indeed found that models performed about 11% to 14% worse on the recreated data, but model *rankings* remained intact. Interestingly, the performance gap narrowed for more recent, higher-accuracy models. For this reason, the authors attribute the observed performance drops not to adaptivity, which should at the very least not decrease over time, but instead to subtle distribution shifts in data collection. We return to this discussion in the context of the extrapolation inference.

The i.i.d. assumption is not directly testable. It serves as a guideline for data collection and puts strong conceptual limitations on the population for which we get valid estimates. ImageNet images were sourced from web search engines and labeled by at least 10 Mechanical Turk workers per image (Deng et al. 2009; Denton et al. 2021). Duplicates and labels on which workers did not agree were filtered out to ensure accurate labels, but this introduces potential sampling biases (Torralba and Efros 2011; Shirali and Hardt 2025), which we discuss further in the conceptualization inference.

[9] Several studies have since examined the empirical impact of this information leakage for other benchmarks (Yadav and Bottou 2019; Roelofs et al. 2019; Miller et al. 2020).

The evaluation set is large enough to yield reliable error estimates with roughly 100 images per class. Russakovsky et al. (2015) empirically confirm this by estimating bootstrap-based confidence intervals, showing low variation in the resulting scores.

*Extrapolation inference* Extrapolation inferences connect expected errors on a specific benchmark with expected errors on alternative benchmarks. In practice, errors often vary substantially across closely related benchmarks (Torralba and Efros 2011; Herrmann et al. 2021, 2024), grounding the need for validity conditions that link ImageNet scores to alternative data distributions, learning tasks, and evaluation metrics. Two sufficient conditions are as follows:

(i) The performance scores from ImageNet are similar across relevant alternative learning problems and data distributions.
(ii) The performance scores for a given evaluation metric are informative about scores from other relevant evaluation metrics.

The two conditions ensure that model performance is *robust* over relevant alternative operationalizations of the task. Robustness is a causal term: The benchmark scores must remain sufficiently stable under changes to a modifier, here the operationalizations (Freiesleben and Grote 2023). What counts as relevant alternatives is determined by the scope of the inference. In our case, this includes essentially *all* alternative operationalizations of the intensional task of image classification. Extrapolation and conceptualization inferences are closely related and can often be treated as a single inferential step. Kane (2013, 28) groups them together, only distinguishing between the respective validity conditions: Extrapolation primarily relies on empirical evidence, whereas conceptualization relies mostly on conceptual or theoretical arguments. Nevertheless, we deem it useful to distinguish between extrapolation and conceptualization inferences, because they serve different interpretative goals and have different targets: The first concerns performance on alternative benchmarks, whereas the second concerns performance on the underlying intensional task.

Let us clarify the difference between the two extrapolation conditions before turning to their evidential support. One might be tempted to unify them into a single requirement of robustness of benchmark scores across alternative learning problems, data distributions, and evaluation metrics. However, robustness across metrics differs from robustness across learning problems and data distributions. Although the latter can be captured using the causal notion of robustness of Freiesleben and Grote (2023), changes to the evaluation metrics will typically change the scores beyond any reasonable tolerance threshold. As such, robustness across metrics either reduces to a weaker requirement, such as the stability of rankings, or must be understood in an information-theoretic sense, where benchmark scores under one metric allow us to infer scores under others.

There is substantial evidence that ImageNet scores robustly *rank* the performance of models on a broad class of image-classification problems, whereas absolute performance scores drop drastically even under small changes. Kornblith et al. (2019) evaluated 16 models trained on ImageNet on alternative image-classification tasks like CIFAR-10/100. The ImageNet scores showed near-perfect rank preservation and were predictive of the performance on the tasks. These findings indicate stronger improvements in cross-task generalization compared with early results by Torralba and Efros (2011). New image datasets like LAIONet (Shirali and Hardt 2025) and ImageNot (Salaudeen and Hardt 2026) provide distributional alternatives to ImageNet. Like Recht et al. (2019), they observe large absolute performance drops but stable rankings compared with ImageNet scores. In particular, Salaudeen and Hardt (2026) compare six eminent learning algorithms that have improved performances on ImageNet and show that they realize similar performance *gains* when trained on ImageNot. Nevertheless, small corruptions or perturbations in the data cause sharp performance drops, raising concerns about the domain on which they are robust (Goodfellow et al. 2015; Geirhos et al. 2018; Hendrycks and Dietterich 2019; Tsipras et al. 2020).

Turning to the evaluation metric, we again find evidence that model rankings are largely consistent, especially across top-1, top-5, and multilabel accuracy (Russakovsky et al. 2015; Shankar et al. 2020). Naturally, the different metrics shift the absolute performance scores. However, when models were evaluated on expert-provided multilabels, allowing for several labels per image, the relative *gains* between the models diminished (Beyer et al. 2020). This indicates metric sensitivity at finer levels of comparison than ordinal rankings.

*Conceptualization inference*. Intended inferences often target intensional tasks rather than specific benchmarks. A conceptualization inference links expected errors on a benchmark to expected errors on a task. Establishing this link requires conditions that connect the extensional description of the task (a benchmark) with its intensional description (Schlangen 2021; Liao et al. 2021). Three sufficient conditions are as follows:

(i) The benchmark captures essential aspects of the theoretical task, for example, image classification.
(ii) The dataset adequately represents the relevant classes.
(iii) The evaluation metric meaningfully reflects predictive performance.

Unlike the former two sub-inferences, the conceptualization inference calls for a more theoretical assessment. The evaluation must address the following questions: Does the learning problem, specified by the input features and target classes of the ImageNet Challenge, adequately capture the relevant aspects of the intensional task of image classification? Does the ImageNet evaluation dataset cover the full breadth of meaning associated with each class? Does the evaluation metric used in the ImageNet Challenge capture what it means to be successful at image classification? We address each of these questions in turn.

Even though the class selection has a clear bias, ImageNet makes a strong effort to capture a wide range of classes relevant to human image classification. The learning problem in ImageNet is to map images, each assigned a single label, to 5 out of 1,000 classes, which aligns with established practices in image classification (Torralba and Efros 2011). The 1,000 classes used in the challenge were first randomly chosen[10] from more than 5,000 classes in the full ImageNet database, which is structured according to the WordNet hierarchy and thus provides a widely accepted ontological basis (Miller 1995; Deng et al. 2009). At the same time, the class selection is very task specific and was later adopted accordingly. For example, 90 classes of different dog breeds were added to increase the between-image difficulty, requiring models to pick up on fine-grained structures (Russakovsky et al. 2015).

The curators explicitly aimed to capture visual diversity along dimensions like size, pose, and lighting (Deng et al. 2009). Nevertheless, the use of an English word database, web-data limitations, image exclusion criteria, annotator priming, and high-agreement filtering during labeling, which can limit diversity, remain as sources of specific task biases (Torralba and Efros 2011; Tsipras et al. 2020; Shirali and Hardt 2025). Probably the most fundamental limitation is that each image only gets a single label, even when multiple may apply. Shankar et al. (2020) estimate that approximately 20% of images of ImageNet have multiple plausible labels. This reflects practical annotation constraints but reduces the benchmark's fidelity to general classification.

The evaluation metric used in the ImageNet Challenge has limited conceptual plausibility. ImageNet Challenge uses the top-5 metric: A prediction is correct if the assigned label is among the model's five guesses. This compensates for the challenge that often multiple plausible labels exist, but it can also lead to distortions. For instance, the prediction {"miniature schnauzer," "tiger," "pirate," "shovel," "fan"} scores perfectly for an image labeled as "miniature schnauzer," despite four very implausible guesses. In contrast, the prediction {"standard schnauzer," "giant schnauzer," "mailbox," "street sign," "Irish terrier"} counts as an error, even though all labels are plausible for an image showing a dog sitting on a street. Multilabel accuracy would better reflect this classification performance and is standard in other benchmarks (Everingham et al. 2009; Everingham et al. 2014).

### 4.3. Constraining the inference

In light of this discussion, one must be cautious when linking benchmark scores on ImageNet to *absolute* expected error rates on image-classification problems more generally. Several of the validity conditions are only partially supported by the available evidence. Although the generalization inference is well supported, adaptivity remains a concern. The extrapolation inference is substantially weakened by the high sensitivity to the specific sampling domain, implying that scores do not reliably transfer across benchmarks. The conceptualization inference is plausible within a certain scope, but the labeling approach and the resulting choice of the top-5 metric introduce important limitations.

Restricting the scope of the interpretation to model *rankings* allows for a more modest and better supported inference. In particular, the extrapolation inference, which is problematic for absolute scores, can be recovered for rankings because models with higher ImageNet performance also tend to perform better on other image-classification benchmarks. The inference based on rankings is therefore more robust and, as Salaudeen and Hardt (2026) argue, sufficient to justify the use of a benchmark like ImageNet for tracking engineering progress in computer vision.

[10] With minor manual postprocessing to exclude obscure classes (Russakovsky et al. 2015).

## 5. Fragile Families Challenge: Are life outcomes predictable?

The FFC, held in 2017, established a predictive benchmark for sociology using data from the Fragile Families and Child Wellbeing Study (Salganik et al. 2019). Predictive benchmarks have the potential to serve as powerful evaluation tools in the quantitative social sciences, complementing explanatory in-sample methods (Verhagen 2022; Shmueli 2010; Hofman et al. 2017). They allow for direct comparisons between different statistical models (Rocca and Yarkoni 2021), testing of model assumptions (Buchholz and Grote 2023), and investigation of theoretical questions like the *predictability* of life outcomes. Results from benchmark challenges like the FFC, therefore, have implications for public policy and are of significance for social theory.

The benchmark challenge was to predict at least one out of six individual life outcomes of 15-year-old teenagers, such as their grade point average (GPA), a psychological construct of perseverance called *grit*, or whether their household experienced material hardship in the past year (Salganik et al. 2019). Built on a 15-year longitudinal survey, more than 4,200 families were interviewed six times. In the end, 160 models were submitted, ranging from conventional statistical methods to various machine learning approaches.

The results, however, were sobering: All submitted models performed poorly in predicting the different outcomes. On the reported evaluation metric, where 1 indicates perfect accuracy and 0 corresponds to predictions no better than the respective average in the training data, even the best models only achieved scores between 0.23 for material hardship, 0.19 for GPA, and 0.03 for lay-off.

### 5.1. The intended inference

Social scientists may aim to infer from the benchmarking challenge that *the best scores achieved in the FFC represent an upper bound on the expected error when predicting individual life outcomes*. If this inference is valid, the FFC results would suggest that life outcomes are inherently difficult to predict—a fundamental limitation rather than an artifact of the survey data or the available modeling techniques.

This inference is very general, giving it an extremely broad scope that spans all life-outcome prediction tasks. The construct of interest is an upper bound on the generalization error. Linking the performance score from the FFC to this bound requires five intermediate steps. First, error rates on the FFC evaluation dataset must generalize to error rates on FFC-like data (generalization inference). Second, they must extend to error rates on alternative life-outcome benchmarks (extrapolation inference). Third, they should capture error rates across life-outcome prediction tasks more broadly (conceptualization inference). Fourth, they must reflect the upper bound of the generalization error for all such tasks (predictability inference). Lastly, we discuss possible consequences from this inference and potential uses (decision inference).

### 5.2. Specifying validity conditions and providing relevant evidence

*Generalization inference*. The generalization inference from the FFC scores is well supported by the careful study design. The final scores are estimated using data from the latest survey wave, which was not publicly available at that time. This ensures independence between the models and the evaluation data.[11] The challenge data were collected as part of the Future Families and Child Wellbeing Study, a representative survey of nonmarital births in large US cities (Reichman et al. 2001). The main threat to generalization lies in the uncertainty in identifying a "winning" model. When bootstrapping several test sets from the evaluation data, the top-ranked model changes in more than half the cases for GPA and nearly a quarter for children's grit (Salganik et al. 2020). Thus, the score rankings partly reflect sampling artifacts rather than genuine performance differences between the models. Nevertheless, changes in the rankings are not a major obstacle to the intended inference because all models performed poorly.

*Extrapolation inference*. The FFC results align with a broad body of research on the predictability of life outcomes, showing that although these outcomes can often be predicted better than chance, overall predictive performance remains low. This empirically supports the extrapolation inference. The FFC alone covers six different prediction targets, already indicating a certain robustness of the results. Despite slightly higher scores for *material hardship* and *GPA*, predictability is low across all targets. Other studies generally find a similar pattern when predicting recidivism, individual-level mortalities, or the risk of long-term unemployment, respectively (Dressel and Farid 2018; Breen and Seltzer 2023; Kuikka 2024; Kern et al. 2021; Desiere and Struyven 2020; Desiere et al. 2019). Zheng and Cheng (2025) report more mixed results specifically for individuals in their 40s: Although *log total income* and *employment status* were poorly predictable, *log hourly wages* reached a substantially higher score of 0.53 on the same metric as used in the FFC. Perdomo et al. (2025) also report high accuracy rates for student dropout prediction. Building on extensive administrative labor and health records from Denmark, Savcisens et al. (2023) recently trained a transformer model to predict individual mortality. Although the model outperforms their simple baseline models, it also only achieves an accuracy of about 78%.

*Conceptualization inference*. The conceptualization inference is generally well supported by the close connection of the FFC to social science theory. The data are drawn from the Fragile Families and Child Wellbeing Study, a well-documented survey with extensive theoretical grounding (Reichman et al. 2001). The prediction targets cover different types of life outcomes, and their operationalizations are established in the field. Nevertheless, life outcomes are a highly heterogeneous type, and the predictability of any specific one may well vary with its scope and specific cultural and societal conditions. Studying the predictability of life outcomes therefore requires making explicit the conditions under which sufficient stability can be expected (Liou et al. 2023; Northcott 2022).

[11] The reported scores may still be overestimated as a result of the finite sample effects: Estimates that correct for this indeed are slightly smaller, but also less precise. Overall, they show patterns similar to those of the original results (Salganik et al. 2020).

A central motivation of the survey underlying the FFC was to examine the role and impact of unwed fathers on child well-being, a group that is often underrepresented in surveys because such fathers are harder to identify (Reichman et al. 2001). This focus on so-called "fragile families" in urban contexts introduces a systematic bias in its data coverage: On average, participating families are poorer and more marginalized than the general US population and face more uncertainties. It is also to be expected that predicting outcomes at the age of 15 is particularly challenging because adolescence is a rather volatile period of life. In contrast, outcomes in later adulthood are considered more predictable because many life trajectories are more stable at this stage (Zheng and Cheng 2025). The evaluation metric underlying the reported normalized out-of-sample $R^2$ scores is the mean squared error (MSE), which disproportionately penalizes large deviations. The choice reflects the research interest of identifying families with "particularly unexpected outcomes" (Salganik et al. 2020).

*Predictability inference.* The intended inference does not only pertain to the prediction errors of a specific set of models but rather to a general limit on the predictability of life outcomes, and thus to a property of the target phenomenon itself. Establishing such a predictability inference requires two validity conditions:

(i) The set of evaluated models must be sufficiently diverse and the data sufficiently large for the best-performing models to approximate the Bayes-optimal model.
(ii) The observed low predictability must not be driven by unmeasured features or the operationalization of the target variable.

The conditions rule out alternative explanations for the low predictive performance observed in the FFC. The first ensures that limited performance cannot be attributed to a restricted set of modeling approaches unable to capture the relationship between features and outcomes or to insufficient data for model training. Under this condition, the best-performing models should approximate the theoretically achievable optimum given the available features and targets, namely, the *Bayes-optimal model* (Hastie et al. 2009). If life outcomes are fundamentally unpredictable, the error of the Bayes-optimal model is high, which means that all predictors perform poorly. However, the Bayes-optimal model may perform poorly for different reasons. The second condition ensures that the poor performance is indeed due to a high irreducible error or high stochasticity in the prediction target, not to uninformative or unmeasured features or a misspecified target variable. With 160 teams independently submitting models based on diverse modeling approaches, the FFC takes a substantial step in this direction. The submissions cover the relevant modeling approaches available at the time of the challenge.[12] Nevertheless, the weak performance of machine learning methods could also be explained by the relatively small size of the training data relative to the large number of features. A total of 4,242 families, split into training, test, and evaluation sets, is a large cohort for a longitudinal survey, considering the costs of a 15-year study. Yet the dataset remains small relative to its 12,942 features. This puts limitations on the intended interpretation as providing bounds on predictability.

[12] Details on all submissions are provided in the appendix of Salganik et al. (2020).

The poor predictive performance may also result from inaccurate or incomplete data, a possibility investigated by Lundberg et al. (2024). To identify limitations that contribute to the prediction errors beyond the sample size, they interviewed 40 children and their families from the FFC, focusing on teenagers whose GPA predictions were inaccurate. From the interviews, they identified three important sources of errors in the FFC: first, imperfectly measured features, such as coarse ones that hide important variation, or recording errors; second, unrecorded features (e.g., one child's higher-than-predicted GPA was explained by a strong social network outside her immediate family, a factor not accounted for in the survey); third, the presence of unmeasurable features, such as events that occurred after the survey was conducted.

*Decision inference*. Validity considerations extend beyond the interpretation of benchmark scores and may also include decision inferences, which combine the scores with a practical *decision rule.* This rule specifies conditions under which a particular course of action is chosen. To illustrate this, model selection based on weather forecasting benchmarks like WeatherBench (Rasp et al. 2020, 2024) is embedded in high-stakes decision-making: Forecasts guide critical decisions, from issuing early warnings for extreme weather to planning energy market operations. Energy providers rely on weather models to forecast solar radiation and wind speeds in order to coordinate renewable energy supply with backup fossil fuel systems. The three conditions for validating such decision inferences, outlined in Kane (2013, 48), are as follows:

(i) The decision rule reflects the utilities of a course of action for the intended goal.
(ii) The decision rule is *fair* across relevant population subgroups.
(iii) The application of the decision rule does not produce undesirable systemic effects compared with alternative rules.[13]

The conditions jointly support the validity of the decision rule and, provided the score interpretation is sound, the resulting decision inference. Importantly, interpreting a score and using it for decision-making are not the same—neither does a valid score interpretation alone justify its use for decision-making, nor does rejecting the application in a given context necessarily undermine the interpretation. Consistent with this distinction, we agree with Kane (2013) that the evaluation of the consequences of both interpreting and using benchmark scores is an essential component of a comprehensive and responsible validity analysis. The consequences of interpreting and using benchmark scores must inform the evidential standards applied.

Predictions of life outcomes similar to those in the FFC regularly inform public policy decisions and the allocation of resources (Perdomo et al. 2025; Zezulka and

[13] For example, in educational settings, standardized testing may influence what is taught in ways that extend beyond its intended role in performance assessment.

Genin 2024). These algorithmic policies often result in differential treatments across social groups (Corbett-Davies et al. 2023; Berk et al. 2023). The three conditions just listed provide general guidelines for validation of such decision inferences.

The intended inference is about a knowledge claim rather than a decision. Nevertheless, such an interpretation can have indirect consequences by influencing future research and policy. Intended and unintended consequences thus arise primarily from the inductive risk associated with the scientific inference drawn (Rudner 1953; Douglas 2000). If we were to mistakenly conclude that life outcomes are only weakly predictable (as the scores suggest), this could impose social costs in the form of less effective and consistent policy interventions (Kleinberg et al. 2017). Conversely, if we erroneously conclude that life outcomes are highly predictable, this could lead to inflated expectations of algorithmic policies, harming potential beneficiaries and misallocating (public) resources. The FFC results thus bear directly on the feasibility and legitimacy of algorithmic governance (Wang et al. 2024). Weighing these competing risks inevitably involves value judgments. Given the careful design of the challenge and the balance of potential costs and benefits, it seems reasonable to treat the results as an important caveat to the development of algorithmic policies and, importantly, to expect low average accuracy and large subgroups for which the models consistently perform poorly.

### 5.3. Constraining the inference

Despite uniformly low scores of all models across the six FFC targets, the results do not support the strong intended inference that life outcomes are *inherently* unpredictable. Although the generalization inference is well supported, the extrapolation inference is less stable because some life outcomes appear to be substantially more predictable. Similarly, the conceptualization inference is fragile, given the heterogeneity of a construct such as "life outcomes." The *predictability* conditions, particularly data coverage and sample size, are especially critical: With larger datasets and more informative features, some life outcomes may, after all, prove to be more predictable.

In light of the validity warrants discussed earlier, a more constrained inference is better supported. The benchmark scores and the validity conditions support the inference that although many individual life outcomes can be predicted better than chance, the overall predictive performance on the data typically available is very likely to be low. This holds particularly for young and marginalized social groups. The challenge also provides evidence that simple models using only a few variables often are competitive with more complex ones.

## 6. Towards a benchmarking epistemology

Predictive benchmarks are powerful methods for the empirical sciences, as the two case studies show. Benchmark scores, together with relevant conditions of validity, can support a wide range of substantial inferences, from tracking research progress in machine learning to assessing the predictability of life events. Drawing on the analogy with psychological testing, we have argued that any such inference requires the explicit specification and evaluation of its validity conditions. The four-step procedure provides a practical guide for interpreting and using benchmark scores.

Analyzing benchmarking as a measurement procedure from a validity-theoretic perspective is valuable beyond these specific inferences. Following, we argue that this framework provides (1) a unifying language for conceptually dispersed discussions in the benchmarking literature, (2) a theoretical grounding for the evaluation of the capabilities of LLMs, (3) insights into the social epistemology of benchmarks, and (4) a clear framework to address the limits of benchmarking.

### 6.1. Unifying language for benchmarking problems

Researchers in machine learning have raised several concerns about the practice of benchmarking. However, the community still lacks a unifying language for articulating problems with benchmarks and the inferences drawn from benchmarking scores. Validity theory provides exactly such a language.

Consider, for example, the influential discussion in computer vision on the problem of *adaptivity*, the sequential reuse of evaluation data over time, which tends to yield inconsistent scores (Blum and Hardt 2015; Yadav and Bottou 2019; Recht et al. 2019). Within validity theory, this is a problem of generalization: Adaptivity violates the second generalization condition—namely, that the evaluation data and the tested model must be independent. The significance of adaptivity becomes clearer from an argument-based perspective because generalization underpins most scientific inferences drawn from benchmark scores. Responses to adaptivity likewise reflect the structure of the argument-based account: Either they weaken the intended inference, shifting from generalization claims about scores to claims about ordinal rankings (Recht et al. 2019; Salaudeen and Hardt 2026), or they propose weaker validity conditions that nonetheless suffice to support the original generalization inference (Dwork et al. 2015; Hardt 2026).

Others have also adopted a validity-theoretic perspective on benchmarks. For instance, Yee (2024) examines *construct legitimacy, criterion validity,* and *construct validity* in evaluating machine learning models in the context of counterterrorism. Recently, Salaudeen et al. (2025) propose a "claim-centred framework" for evaluating general-purpose AI systems that is closely related to our approach. The attribution of *capabilities* to AI systems on the basis of benchmark scores has spurred growing interest in psychometric testing and validity (Wang et al. 2026; Sühr et al. 2025; Ye et al. 2025).

### 6.2. Guiding LLM capability testing

With the development of generative AI and LLMs, the role of benchmarks in machine learning has changed. Although these models are typically trained as next-token predictors, some researchers aim to attribute *capabilities* to them and have begun to develop benchmarks to draw inferences about such latent constructs. For example, benchmarks have been designed to test whether models have capabilities like a stable sense of morality (Jiang et al. 2021), a theory of mind (Strachan et al. 2024), or legal reasoning (Masry et al. 2022). Attributing such capabilities to LLMs has been widely criticized both for anthropomorphizing machine learning models (Shanahan 2024; Sypherd et al. 2025) and particularly for lacking a sound methodology (Shapira et al. 2024; Zhuang et al. 2025).

Although we agree that the anthropomorphization of machine learning tools carries risks, we also believe that LLM capability tests can serve an important inferential role. LLMs are assessed as general-purpose models; however, users are typically concerned with their usefulness for specific downstream tasks. Thus, capability scores must provide evaluation at a level of abstraction that is informative for a wide range of tasks while remaining feasible to implement. This makes the lack of methodological standards highly problematic. Most of the targeted capabilities cannot be straightforwardly operationalized in a benchmark; they are what Kane (2013) calls *theoretical properties*, instead of *measurable attributes*.[14] Validity theory in turn was explicitly developed to infer latent constructs from measurable test scores. For this reason, it can guide LLM capability testing in establishing evaluation and inferential standards.

Cronbach and Meehl (1955), Whitely (1983), and Borsboom et al. (2003) have argued for different approaches to developing theories about latent constructs, using nomological networks, process theories, or structural causal models, respectively. The argument-based account of validity also allows for the integration of such theories and supports inferences about latent constructs. The validity conditions of such *theorization inferences* from Kane (2013) can be fruitfully adopted when reasoning about LLM capabilities. In addition to the generalization, extrapolation, and conceptualization conditions, they require (i) a theory that explicates the latent construct, explains its relationship to other such constructs, and provides a justification of the plausibility of the theory, and (ii) evaluation data that includes samples representing the measurable attributes that the theory links to the construct of interest.

Decades of theorizing human cognition demonstrate that such conditions are more easily stated than met (McGrew 2005). First, it remains elusive how LLM "cognition" operates and whether it can be meaningfully compared to human cognition, making such transfers particularly challenging (Shanahan 2024). Second, even if suitable theoretical models were available, they would need to be rigorously tested before supporting any inference. In particular, they should undergo *convergent and discriminant analyses* to ensure that the theory coheres with empirical results (Cronbach and Meehl 1955; Messick 1989; Strauss and Smith 2009). Research on human testing highlights the difficulties in designing adequate assessment tests that are broad enough to capture the behavioral implications of a latent construct without becoming too encompassing. Tests are constantly at risk of either failing to represent certain aspects of the construct, called *construct underrepresentation*, or being influenced by factors irrelevant to the construct, referred to as *construct irrelevance* (Strauss and Smith 2009). Construct irrelevance, in particular, is difficult to detect, because depending on the external factors included, it may either facilitate or hinder test performance (Messick 1989).

In agreement with Sühr et al. (2025), we believe that LLM researchers should not simply apply tests designed for humans to LLMs. At the same time, we are also convinced that methodological transfers like the one we propose here from psychological testing to assessing LLM capabilities can be very fruitful.

---

[14] Raji et al. (2021) argue that only well-specified intensional tasks can be faithfully operationalized by benchmarks. Many general-purpose language benchmarks, such as SuperGLUE, may be seen to violate this principle (Dorner et al. 2023; Millière and Buckner 2024; Zhang and Hardt 2024; Grote et al. 2024a).

### 6.3. Social dimensions of benchmarking

So far, we used validity theory to analyze the epistemic role of predictive benchmarks in drawing inferences, largely leaving aside their important social role in structuring research communities. How do benchmarks shape research practices and communities? And who holds the power to establish benchmarks within a community?

These questions have begun to receive some attention. Raising concerns about biased data and task selection, Koch et al. (2021) show that research is increasingly concentrated on a small set of benchmark datasets, often repurposed from other contexts and provided by a few powerful institutions. Dehghani et al. (2021) argue that when a single benchmark dominates a field of research, this favors methods that are already well aligned with the benchmark's idiosyncrasies. Conversely, where multiple benchmarks coexist, researchers may cherry-pick those on which their models happen to perform best. These results point to a fragility of benchmark-based evaluation, which is especially serious when model assessment is reduced to benchmarking. Hooker (1995) further cautions that framing scientific research as a competition promotes a race for marginal numerical improvements at the expense of more substantive progress.

Although validity theory cannot replace the development of a substantial social epistemology of benchmarking, it draws attention to the relations between the curators of a benchmark and those intending to draw inferences from it. Typically, machine learning practitioners and scientists employ benchmarks they did not curate themselves. As a consequence, many of the validity conditions discussed earlier, including generalization and extrapolation conditions, are outside their control. This underscores the need for *benchmark curators* to maintain close communication with benchmark users, ensuring that benchmarks are designed to support relevant inferences and that underlying design choices are clearly communicated in the corresponding publications. The social dimension of benchmarking highlights challenges for drawing valid inference from benchmark scores. Any comprehensive validity theory of benchmarking will have to address these challenges in future work.

### 6.4. Limits of the benchmarking epistemology

Predictive benchmarks were originally developed to shift the focus of evaluation away from broad claims about artificial general intelligence and toward measuring performance on concrete and narrowly defined learning problems. Ironically, with the advent of modern LLMs, researchers have returned to making such claims about general capabilities—this time on the basis of benchmark scores. This raises the question of the limits of benchmarks: What can be meaningfully measured by and inferred from them, and what cannot?

Validity theory, in particular the argument-based account, teaches an important lesson here: The more abstract and complex the intended inferences become, the stronger must the corresponding validity conditions in the interpretation or use argument be (Kane 2013). Stronger validity conditions impose higher demands on supporting evidence while also creating more opportunities for potential undermining evidence. As a result, such inferences will often fail because some of the validity conditions are only partially supported or even undermined. For strong theoretical inferences—which are, of course, of particular interest in the empirical sciences—these failure modes can often be traced to benchmarks "thinning the world" by focusing on easily quantifiable aspects of a phenomenon, under-representing the construct of interest, or capturing irrelevant features of it (Raji et al. 2021; Denton et al. 2021; Orr and Kang 2024). This danger becomes particularly acute when benchmarks come to function as authoritative proxies for entire tasks, as ImageNet once did for image classification, gaining legitimacy through widespread adoption within the research community (Orr and Crawford 2024).

On first sight, benchmarking may appear to be the tool to establish the mystical ideal of a *theory-free science*—a science devoid of human theoretical preconceptions and shaped only by empirical data and predictions (Andrews 2025). However, the critical evaluation of benchmarking as a measurement procedure demonstrates the importance of a validity theory and makes clear that theory, for example, in the form of validity conditions, remains indispensable for drawing inferences from benchmark scores. In other words, predictive benchmarking is a key site of conceptual and theoretical reasoning in machine learning, and the limits of the benchmarking epistemology are not determined by the benchmarks themselves but by the strength and justification of the inferences we draw from them.

**Acknowledgments.** We thank Konstantin Genin, Jan-Willem Romeijn, Wolfgang Spohn, Bob Williamson, Tom Sterkenburg, Thomas Grote, Markus Ahlers, Katharina Holzhey, Kino Zhao, Moritz Hardt, Luis Winckelmann, Tim Schreier, and Alex Braun for helpful discussions and valuable feedback. We are also grateful to the two anonymous reviewers for their careful reading and insightful comments, which substantially improved the article.

**Funding Statement.** This work has been funded by the Deutsche Forschungsgemeinschaft (DFG; German Research Foundation) through the Emmy Noether Research Group "From Bias to Knowledge: The Epistemology of Machine Learning" (project number 511917847) and under Germany's Excellence Strategy, EXC 2064/1 (project number 390727645). The authors acknowledge support for Timo Freiesleben from the Carl Zeiss Stiftung (Project: "Certification and Foundations of Safe Machine Learning Systems in Healthcare") and the Munich Center for Machine Learning. They also thank the International Max Planck Research School for Intelligent Systems (IMPRS-IS) for supporting Sebastian Zezulka.

**Declarations.** None to declare.